\documentclass[letterpaper, 10pt, conference]{ieeeconf}  

\IEEEoverridecommandlockouts                             

\usepackage[pdftex]{graphicx}
\usepackage{cite}
\usepackage{verbatim}		
\usepackage{amsmath}
\usepackage{amssymb}

\usepackage{amsthm}
\usepackage{mathtools}	
\usepackage{grffile}	
\usepackage[tight,footnotesize]{subfigure}
\usepackage{microtype} 
\usepackage{color}
\usepackage{url}
\usepackage[ruled,vlined,linesnumbered]{algorithm2e}
\usepackage{bm} 
\usepackage{comment}
\usepackage{arydshln}
\let\labelindent\relax
\usepackage{enumitem}

\usepackage{adjustbox}
\usepackage{tikz}
\usetikzlibrary{shadings, patterns, angles, quotes, arrows.meta, shapes, decorations.pathmorphing, decorations.shapes, decorations.text}
\usetikzlibrary{calc,intersections,arrows.meta}
\usepackage{pgfplots}

\usepackage{hyperref}
\hypersetup{
	colorlinks=true,
	linkcolor=blue,
	citecolor=red,
	urlcolor=black,
}

	\tikzset{
	pil/.style={
		->,
		thick,
		shorten <=2pt,
		shorten >=2pt,}
}

\theoremstyle{remark}
\newtheorem{remarkx}{Remark}
\newenvironment{remark}
{\pushQED{\qed}\remarkx}
{\popQED\endremarkx}

\theoremstyle{definition}

\newtheorem{assump}{Assumption}
\newtheorem{problem}{Problem}
\newtheorem*{problem*}{Problem}

\theoremstyle{plain}
\newtheorem{theorem}{Theorem}

\newcommand{\defeq}{:=} %
\newcommand{\hgh}[1]{{#1}^{\mathrm{hgh}}}
\newcommand{\phy}[1]{{#1}^{\mathrm{phy}}}
\newcommand{\matr}[1]{\begin{bmatrix} #1 \end{bmatrix}}
\newcommand{\transpose}[1]{#1^\top}
\newcommand{\diag}[1]{{\rm diag}\{ #1\}}
\newcommand{\chiup}{\raisebox{2pt}{$\chi$}}
\newcommand{\vf}{\chiup}
\newcommand{\set}[1]{\mathcal{#1}}

\newcommand{\norm}[1]{\left\lVert#1\right\rVert}
\newcommand{\mbr}[1][{}]{\mathbb{R}^{#1}}	%

\newcommand{\vfpf}{\prescript{\rm pf}{}{\vf}}
\newcommand{\vfco}{\prescript{\rm co}{}{\vf}}
\newcommand{\vfcb}{\mathfrak{X}}%
\newcommand{\normv}[1]{\overline{#1}}			%

\newcommand{\multivfcb}[2]{\vfcb_{#1#2}}
\newcommand{\multihatvfcb}[2]{ {\normv{\vfcb_{#1}}}_{#2} }

\newcommand{\multiphi}[2]{\phi_{#1#2}}
\newcommand{\multif}[2]{f_{#1#2}}
\newcommand{\multifdot}[2]{f_{#1#2}'}

\newcommand{\multix}[2]{x_{#1#2}}
\newcommand{\multik}[2]{k_{#1#2}}

\newcommand{\sat}{\mathrm{Sat}_{a}^{b}}

\newcommand{\scalemath}[2]{\scalebox{#1}{\mbox{\ensuremath{\displaystyle #2}}}}

\title{\LARGE \bf
Distributed coordinated path following using guiding vector fields
}

\author{Weijia Yao, Hector Garcia de Marina, Zhiyong Sun, Ming Cao %
	\thanks{Weijia Yao and Ming Cao are with the University of Groningen, the Netherlands. {\tt\small \{w.yao,m.cao\}@rug.nl} Hector Garcia de Marina is with Universidad Complutense de Madrid, Spain. {\tt\small hgarciad@ucm.es} Zhiyong Sun is with the Eindhoven University of Technology, the Netherlands. {\tt\small z.sun@tue.nl} The work of Cao was supported in part by the European Research Council (ERC-CoG-771687) and the Netherlands Organization for Scientific Research (NWO-vidi-14134). The work of Yao is supported in part by the China Scholarship Council. The work of Hector Garcia de Marina is supported by the grant Atraccion de Talento 2019-T2/TIC-13503 from the Government of the Autonomous Community of Madrid.
	}%
}

\begin{document}

\maketitle
\thispagestyle{empty}
\pagestyle{empty}

\begin{abstract}
	It is essential in many applications to impose a scalable coordinated motion control on a large group of mobile robots, which is efficient in tasks requiring repetitive execution, such as environmental monitoring. In this paper, we design a guiding vector field to guide multiple robots to follow possibly different desired paths while coordinating their motions. The vector field uses a path parameter as a virtual coordinate that is communicated among neighboring robots. Then, the virtual coordinate is utilized to control the relative parametric displacement between robots along the paths. This enables us to design a saturated control algorithm for a Dubins-car-like model. The algorithm is distributed, scalable, and applicable for any smooth paths in an $n$-dimensional configuration space, and global convergence is guaranteed. Simulations with up to fifty robots and outdoor experiments with fixed-wing aircraft validate the theoretical results.
\end{abstract}

\section{Introduction}

One of the grand challenges in robotics is the reliable and systematic coordination of a large number of robots without a central nexus \cite{yang2018grand}. Tasks involving vast geographic areas or volumes, such as search and rescue, patrolling, and environmental monitoring, are practically impossible for a single robot, while multiple coordinated robots can deal with such scenarios efficiently. In these applications, it is fundamental for a group of robots to accurately follow possibly different paths, while coordinating their motions to form a formation which satisfies some geometric or parametric constraints. In this paper, we design a \emph{coordinating vector field} to guide multiple robots to converge to their respective desired paths and realize the aforementioned coordinated behavior in a distributed way through local information exchange. %

\textbf{Related work:}
Many path-following algorithms for a single robot have been studied in the literature, such as Line-of-sight (LOS) \cite{fossen2003line} among others \cite{Sujit2014}. Among several tested algorithms, those using a guiding vector field are shown to achieve the highest path-following accuracy while requiring the least control efforts \cite{Sujit2014}. However, to the best of our knowledge, there are only a few studies on extending the guiding vector field approach to realize multi-robot coordinated path following tasks. Utilizing the guiding vector field in \cite{kapitanyuk2017guiding}, the work \cite{de2017circular} employs a distributed algorithm to change the radii of the circular paths for each fixed-wing aircraft flying at a constant speed such that the aircraft asymptotically follow a circular path of the same radius while keeping user-specified distances. Another work \cite{nakai2013vector} designs different potential functions to derive a guiding vector field for multiple robots, but their motions cannot be explicitly coordinated (e.g., the multi-robot distance cannot be controlled). These two studies only consider some common desired paths, such as a circle or a straight line in 2D. The paper \cite{pimenta2013decentralized} provides a decentralized control law such that multiple robots can circulate a closed curve in 3D, but the curve must have a specific form, and some strong assumptions on the integral curves are required. 

There are many studies on coordinated path-following \emph{without} employing a guiding vector field. The robots' motions in \cite{ghommam2010formation} are coordinated by using a virtual structure when following predefined paths. However, each robot needs to broadcast its states and reference trajectories to the rest of the robots, so the communication burden increases rapidly as the size of the multi-robot systems increases. Recently, \cite{chen2019coordinated} proposes a distributed hybrid control law based on \cite{lan2011synthesis} to coordinate the robot motions such that robots keep a constant parametric separation. A similar task is achieved by another recent work in \cite{wang2019cooperative}. However, these two studies, as well as many others (e.g., \cite{burger2009straight,doosthoseini2015coordinated,reyes2015flocking}), are restricted to planar or simple desired paths, such as a straight line. Another related work \cite{zhang2007coordinated} only considers planar simple closed curves and unit-speed particles, and the convergence result is valid locally. In \cite{sabattini2015implementation}, an output-regulation controller is applied to certain robots while other robots are controlled through local interactions. This approach is effective, but the desired motion must be periodic and generated by a linear exosystem.  

Multi-robot coordinated path following control has been extensively studied in the specialized cases of circular formation control and circumnavigation control. In circular formation control, multiple robots either need to move to or initially start on a specified circle, and then move along the circle until some desired arc distances are achieved via local communication \cite{iqbal2019circular,st2018circle,zhu2014cooperative,sun2018circular,wang2014controlling,wang2013forming,sepulchre2007stabilization}. Circumnavigation control additionally requires all robots to keep moving along the circle to fulfill some tasks, such as escorting an important target \cite{fonseca2019cooperative,shames2011circumnavigation,deghat2014localization,zheng2014circumnavigation,yao2019distributed,yao2017distributed,yao2018distributed,ma2018cooperative,lu2019distributed}. However, these studies are all restricted to 2D circular paths, and many of them impose additional requirements on the robots' initial positions (e.g., to maintain the order preservation \cite{wang2013forming,wang2014controlling}). The paper \cite{mong2006pattern} presents a decentralized controller for robots characterized by the double-integrator model  such that they can generate 2D desired geometric patterns composed of simple closed curves. This work is extended in \cite{mong2007stabilization} for robots described by the single-integrator model to circulate a closed curve. However, these two studies are both restricted to 2D closed curves.

\textbf{Contributions:}
We propose a coordinating vector field to enable distributed coordinated path following for multiple robots, and a corresponding saturated control algorithm is designed for a Dubins-car-like model. In particular, we use the path parameter as an additional virtual coordinate for the guiding vector field, and this virtual coordinate is further communicated among neighboring robots to achieve motion coordination. There are many appealing features of our approach: 1) It is distributed and scalable, and thus applicable for a team of an arbitrary number of robots; 2) It can deal with complex paths such as self-intersecting, non-convex, or non-closed ones in an $n$-dimensional configuration space. 3) Given a fixed communication frequency, the communication burden is low since every two neighboring robots only need to transmit and receive at most two scalars (i.e., the virtual coordinate and its time derivative); 4) It guarantees that the coordinated path-following control objectives are achieved globally regardless of robots' initial positions. 5) The approach has been verified by simulations with a large multi-robot system and by experiments with fixed-wing aircraft in a windy environment. We also show that our approach is promising in area or volume coverage tasks by exploiting a 2D or 3D Lissajous curve with \emph{irrational coefficients}, as it is guaranteed that robots can eventually visit every point of an area or volume while keeping prescribed distances.

\textbf{Notations:} The notation $\mathbb{Z}_{i}^{j}$ represents the set $\{m \in \mathbb{Z} : i \le m \le j \}$. We use boldface for a vector $\bm{v} \in \mbr[n]$, and its $i$-th entry is denoted by $v_i$ for $i\in \mathbb{Z}_{1}^{n}$. Consider a system consisting of $N$ robots, and a vector $\bm{u_i} \in \mbr[n]$ for $i \in \mathbb{Z}_{1}^{N}$, then for $j \in \mathbb{Z}_1^n$, the notation $u_{ij}$ refers to the $j$-th entry of $\bm{u_i}$. {\bf Graphs:} We call $\set{V} := \{1,\dots,N\}$ the \emph{node set} representing the robots, and $\set{E}\subseteq(\mathcal{V}\times\mathcal{V})$ is the \emph{edge set} encoding the communication links. We call $\mathcal{N}_i:=\{j\in\mathcal{V}:(i,j)\in\mathcal{E}\}$ the set of neighbors of robot $i$. In this paper, we only consider undirected graphs, which implies that if $(i,j) \in \set{E}$, then Robot $i$ and Robot $j$ can share information bidirectionally. Please see \cite{mesbahi2010graph} for an introduction to graph theory.

\section{Preliminaries}

Guiding vector fields with virtual coordinates for robot navigation enjoy many desirable features, such as guaranteeing global convergence to the desired path and enabling self-intersecting desired path following \cite{yao2021singularity}. We present here a brief introduction. Suppose the desired path $\phy{\set{P}}$ is parameterized by
$
x_{1} = f_1(w), \dots, x_{n}=f_n(w),
$
where $x_i$ is the $i$-th coordinate, $w \in \mbr[]$ is the parameter of the path and $f_i$ is twice continuously differentiable (i.e., $f_i \in C^2$), for $i\in \mathbb{Z}_{1}^{n}$. To derive a corresponding guiding vector field, one needs to describe the desired path as the intersection of $n$ hyper-surfaces \cite{yao2018cdc,Goncalves2010,lawrence2008lyapunov,yao2020auto,yao2020topo}. To this end, taking $w$ as an additional argument, one defines $n$ \emph{surface functions} $\phi_i: \mbr[n+1] \to \mbr$ as follows: $\phi_i(\bm{\xi})=x_{i}-f_i(w)$ for $i \in \mathbb{Z}_1^n$, where $\bm{\xi}=(x_1,\dots,x_n,w) \in \mbr[n+1]$ is the generalized coordinate with an additional (virtual) coordinate $w$. Therefore, the desired path with an additional coordinate is the intersection of the hyper-surfaces described by the zero-level set of these functions; i.e., $\hgh{\set{P}}\defeq \{ \bm{\xi} \in \mbr[n+1]: \phi_i(\bm{\xi})=0, i\in \mathbb{Z}_{1}^{n} \} $. The projection of $\hgh{\set{P}}$ on the plane spanned by the first $n$ coordinates is the original desired path $\phy{\set{P}}$, so we can use the higher-dimensional guiding vector field $\vf: \mbr[n+1] \to \mbr[n+1]$ corresponding to $\hgh{\set{P}}$ to make a robot follow the original desired path $\phy{\set{P}}$ by projecting to the $n$-dimensional space \cite{yao2021singularity}, where
\begin{equation} \label{eq_gvf}
\scalemath{1}{
\vf(\bm{\xi}) = \times \big( \nabla \phi_i,\dots, \nabla \phi_n \big) - \sum\nolimits_{i=1}^{n} k_i \phi_i \nabla \phi_i,
}
\end{equation}
with $k_i>0$,  $\nabla \phi_i \in \mbr[n+1]$ being the gradient of $\phi_i$ with respect to the generalized coordinate $\bm{\xi}$, and $\times(\cdot)$ being the wedge product \cite[Chapter 7.2]{galbis2012vector}. The physical interpretation of the vector field $\vf$ is clear: the second term $ - \sum_{i=1}^{n} k_i \phi_i \nabla \phi_i$ is a weighted sum of all the gradients, which guides the robot trajectory towards the intersection of the hyper-surfaces (i.e., the desired path), while the first term $\times \big( \nabla \phi_i,\dots, \nabla \phi_n \big)$, being orthogonal to all the gradients \cite[Proposition 7.2.1]{galbis2012vector}, provides a propagation direction along the desired path.

In our previous work \cite{yao2021singularity,yao2020mobile}, this higher-dimensional guiding vector field in \eqref{eq_gvf} is proved to possess no singular points, where the vector field becomes zero, thanks to the additional dimension (i.e., the additional virtual coordinate $w$). However, like many other current studies (e.g., \cite{Goncalves2010,lawrence2008lyapunov,nelson2007vector}), the vector field is used for the guidance of only one single robot. We aim to extend the vector field to include a coordination component and achieve motion coordination among multiple robots. Since the additional virtual coordinate $w$ not only helps eliminate singular points, but also acts as the path parameter, one idea is to utilize the virtual coordinate $w$ to coordinate the robots' motions. Thus, the question is:

\textbf{Question:} Suppose there are $N$ robots. Based on the higher-dimensional vector field in \eqref{eq_gvf}, how can one design an extra \emph{coordination mechanism} using the virtual coordinates $w_i$, $i\in \mathbb{Z}_{1}^{N}$, such that 1) Each robot can follow their desired paths; 2) All robots can coordinate their motions by controlling the virtual coordinates $w_i,i\in \mathbb{Z}_{1}^{N}$ via local information exchange among neighbors?%

The extra coordination mechanism and the precise meaning of motion coordination will become clear in the next section, where a mathematical problem formulation based on dynamical systems theory will be presented.

\section{Design of guiding vector fields for multi-robot systems} \label{sec: gvf}

\subsection{Mathematical problem formulation} \label{sec: gvf1}
The $i$-th robot is required to follow a path in $\mbr[n]$, parameterized by $n$ parametric equations:
\begin{equation} \label{eq_param_func}
	\multix{i}{1} = \multif{i}{1}(w_i)	\quad \dots  \quad 	 \multix{i}{n} = \multif{i}{n}(w_i),
\end{equation}
where $\multix{i}{j}$ is the $j$-th coordinate, $\multif{i}{j} \in C^2$ is the $j$-th parametric function for the $i$-th robot, for $i\in \mathbb{Z}_{1}^{N}, j\in \mathbb{Z}_{1}^{n}$, and $w_i \in \mbr[]$ is the parameter of the desired path. To derive the path-following guiding vector field, we use the parameter $w_i$ as an additional virtual coordinate, and the higher-dimensional desired path is described by
\[
	\set{P}_i \defeq \{ \bm{\xi_i} \in \mbr[n+1] : \multiphi{i}{1}(\bm{\xi_i})=0, \dots,  \multiphi{i}{n}(\bm{\xi_i})=0 \},
\]
where $\bm{\xi_i} \defeq (\multix{i}{1}, \dots, \multix{i}{n}, w_i) \in \mbr[n+1]$ denotes the \emph{generalized coordinate} of the $i$-th robot, and the surface functions are $\multiphi{i}{j}=\multix{i}{j}-\multif{i}{j}(w_i)$ for $i\in \mathbb{Z}_{1}^{N}$ and $j\in \mathbb{Z}_{1}^{n}$. Define $\bm{\Phi_i}(\bm{\xi_i}) \defeq \transpose{(\multiphi{i}{1}(\bm{\xi_i}), \dots, \multiphi{i}{n}(\bm{\xi_i}))} \in \mbr[n]$. Observe that $\bm{\xi}_i \in \set{P}_i$ if and only if $\norm{\bm{\Phi_i}(\bm{\xi}_i)}=0$. Therefore, one can use $\bm{\Phi_i}(\bm{\xi_i})$ to quantify the distance to the desired path $\set{P}_i$; we call $\bm{\Phi_i}(\bm{\xi_i})$ the \emph{path-following error} to $\set{P}_i$. The aim is to design guiding controllers such that the norm $\norm{\bm{\Phi_i}(\bm{\xi_i})}$ converge to zero eventually. By combining \eqref{eq_gvf} and \eqref{eq_param_func} , we obtain the analytic expression of the \emph{path-following guiding vector field} $\bm{\vfpf_i}: \mbr[n+1] \to \mbr[n+1]$ for the $i$-th robot:
\begin{align} \label{eq_vfpf}
\scalemath{1}{
\bm{\vfpf_i}(\bm{\xi_i}) = \matr{ (-1)^n \multifdot{i}{1}(w_i) - \multik{i}{1} \multiphi{i}{1}(\bm{\xi_i}) \\  
					\vdots \\ 
					(-1)^n \multifdot{i}{n}(w_i) - \multik{i}{n} \multiphi{i}{n}(\bm{\xi_i}) \\ 
					(-1)^n + \sum_{l=1}^{n} \multik{i}{l} \multiphi{i}{l}(\bm{\xi_i}) \multifdot{i}{l}(w_i) } %
}
\end{align}
for $i\in \mathbb{Z}_{1}^{N}$, where $k_{ij}>0$ are constant gains, and $f'_{ij}$ are the derivatives of $f_{ij}$ with respect to the argument $w_i$.%

To achieve coordination in $w_i$, thus indirectly coordinate the positions of robots, we introduce the \emph{coordination component} $\bm{\vfco_i}: \mbr[N] \to \mbr[n+1]$ for the $i$-th robot, $i\in \mathbb{Z}_{1}^{N}$, which is defined to be 
\begin{align} \label{eq_vfco}
\bm{\vfco_{i}}(\bm{w}) = \transpose{\big( 0, \cdots, 0, c_i(\bm{w}) \big)}, 
\end{align}
where $\bm{w}=\transpose{(w_1, \dots, w_N)}$ and $c_i: \mbr[N] \to \mbr$ is called the \emph{coordination function} to be designed later to enable coordination among robots through local interactions via the neighboring virtual coordinates $w_j$ for $j \in \set{N}_i$.  Specifically, the virtual coordinates of multiple robots $w_i(t) - w_j(t)$ are to be steered to converge to the \emph{desired relative states} $\Delta_{ij} \in \mbr$ for $(i,j) \in \set{E}$. We will see how to design $\Delta_{ij}$ in the beginning of Section \ref{sec: gvf2}.

We design the $i$-th \emph{coordinating vector field} $\bm{\vfcb_i}: \mbr[n+N] \to \mbr[n+1]$ to be the weighted sum of the path-following vector field $\bm{\vfpf_i}$ and the coordination component $\bm{\vfco_i}$ as below:
\begin{align}	\label{eq_vf_combined}
	\bm{\vfcb_i}(\bm{\xi_i}, \bm{w}) = \bm{\vfpf_i}(\bm{\xi_i}) + k_c \bm{\vfco_i}(\bm{w}),
\end{align}
where $k_c>0$ is a parameter to adjust the weighted contribution of $\bm{\vfpf_i}$ and  $\bm{\vfco_i}$ to $\bm{\vfcb_i}$. With a larger value of $k_c$, the motion coordination is achieved faster. The coordinating vector field $\bm{\vfcb_i}$ represents the desired moving direction for a robot, so the robot is guided. Thus it is key to study the guidance result, or precisely, the convergence results of the integral curves of the vector field $\bm{\vfcb_i}$ for $i\in \mathbb{Z}_{1}^{N}$. Precisely, we define $\bm{\xi} \defeq \transpose{(\transpose{\bm{\xi_i}}, \dots, \transpose{\bm{\xi_N}})} \in \mbr[(n+1)N]$ and $\bm{\vfcb}(\bm{\xi}) \defeq \transpose{(\bm{\vfcb_1},\dots,\bm{\vfcb_N})} \in \mbr[(n+1)N]$, and study the integral curves of $\bm{\vfcb}(\bm{\xi})$; i.e., the trajectories to the differential equation 
\begin{equation} \label{eq_ode}
	\dot{\bm{\xi}} = \bm{\vfcb}(\bm{\xi}),
\end{equation}
given an initial condition $\bm{\xi_0} \in \mbr[(n+1)N]$ at $t=0$. Now we can formally formulate the problem as follows:
\begin{problem} \label{problem1}
	Design the coordinating vector field $\bm{\vfcb_i}$ in \eqref{eq_vf_combined} for $i\in \mathbb{Z}_{1}^{N}$, such that the trajectories of \eqref{eq_ode}, given an initial condition $\bm{\xi_0} \in \mbr[(n+1)N]$, fulfill the two control objectives: 
	\begin{enumerate}[leftmargin=*]
		\item (\textbf{Path Following}) Robot $i$'s path following error to its desired path $\set{P}_i$ converges to zero asymptotically for all $i\in \mathbb{Z}_{1}^{N}$. That is, $\norm{\bm{\Phi_i}(\bm{\xi_i}(t)} \to 0$ as $t \to \infty$, $\forall i \in \mathbb{Z}_{1}^{N}$.
		\item (\textbf{Motion Coordination}) Each robot's motion is coordinated distributedly subject to the communication graph $\set{G}=(\set{V}, \set{E})$ such that their additional virtual coordinates satisfy $w_i(t) - w_j(t) - \Delta_{ij} \to 0$  as $t \to \infty$ for $(i,j) \in \set{E}$.
	\end{enumerate}
\end{problem}
We require two mild standing assumptions that are reasonable in practical applications:
\begin{assump} \label{assump_graph}
	The communication graph $\set{G}=(\set{V}, \set{E})$ is connected.
\end{assump}
\begin{assump} \label{assump_bounded}
	The first and second derivatives of $f_{ij}(\cdot)$ are bounded for $i\in \mathbb{Z}_{1}^{N}, j\in \mathbb{Z}_{1}^{n}$. %
\end{assump}

\subsection{Design of the coordination component}  \label{sec: gvf2}
Given the desired path $\mathcal{P}_i$, we can design the desired relative states $\Delta_{ij}$ starting from a particular reference configuration $\bm{w}^* := \transpose{(w^*_1, \cdots, w^*_N)}$. Hence, $\bm{\Delta}^* = D^\top \bm{w}^*$ is the stacked vector of $\Delta_{ij}, (i,j)\in\mathcal{E}$ \cite{garciademarina2020maneuvering}, where $D \in \mbr[N \times |\set{E}|]$ is the incidence matrix \cite[p. 23]{mesbahi2010graph}.
Now we propose to employ the following consensus control algorithm \cite[p. 25]{ren2008distributed}:
\begin{equation} \label{eq_coordination}
	c_i =  - \sum_{j\in\mathcal{N}_i} \big( w_i - w_j - \Delta_{ij} \big), \; \forall i \in \mathbb{Z}_{1}^{N}.
\end{equation}
Equation (\ref{eq_coordination}) can be written in a compact form as
\begin{equation} \label{eq_coordination_stack}
	\bm{c}(\bm{w}) = -L (\bm{w} - \bm{w}^*) = -L \bm{\tilde w},
\end{equation}
where $\bm{c}(\bm{w})=\transpose{(c_1(\bm{w}),\dots,c_N(\bm{w}))}$, $L=L(\set{G})$ is the Laplacian matrix and $\bm{\tilde{w}} \defeq \bm{w} - \bm{w}^*$. Combining \eqref{eq_vfpf}, \eqref{eq_vfco}, \eqref{eq_vf_combined} and \eqref{eq_coordination}, we have attained the coordinating vector field $\bm{\vfcb_i}$ for $i\in \mathbb{Z}_{1}^{N}$. %
\begin{remark}
From \eqref{eq_vfco}, \eqref{eq_vf_combined} and \eqref{eq_coordination}, one observes that neighboring information exchange only happens in the coordination component $c_i$. Notably, the communication burden is low: Robot $i$ transmits only a scalar $w_i$ to the neighboring Robot $j \in \set{N}_i$.
\end{remark}
\subsection{Convergence analysis}  \label{sec: gvf3}
The convergence analysis of trajectories to \eqref{eq_ode} is nontrivial given that the right-hand side of \eqref{eq_ode} is \emph{not} a gradient of any potential function, since the path-following vector field in \eqref{eq_gvf} contains a wedge product of all the gradients. In this subsection, we show that the coordinating vector field \eqref{eq_vf_combined} enables multiple robots to follow their desired paths while being coordinated by the virtual coordinates such that $w_i(t) - w_j(t)$ converge to $\Delta_{ij}$ for $(i,j) \in \set{E}$ as $t\to\infty$.%

For simplicity, we first consider Robot $i$, and most of the function arguments are ignored henceforth unless ambiguity arises. Define $K_i \defeq \diag{\multik{i}{1}, \dots, \multik{i}{n}}$ and $\bm{f'_i}(w_i) \defeq \transpose{(\multifdot{i}{1}(w_i), \dots, \multifdot{i}{n}(w_i) )}$. Then one can calculate that $\scalemath{0.9}{ \transpose{ (\transpose{\nabla \multiphi{i}{1}}  \bm{\vfpf_i},\cdots, \transpose{\nabla \multiphi{i}{n}}  \bm{\vfpf_i} )}  = - K_i \bm{\Phi_i}  - \bm{f'_i}(w_i) \transpose{\bm{f'_i}(w_i)} K_i \bm{\Phi_i} } $ and $\scalemath{0.9}{ \transpose{ (\transpose{\nabla \multiphi{i}{1}}   \bm{\vfco_i}(\bm{w}), \cdots,  \transpose{\nabla \multiphi{i}{n}}  \bm{\vfco_i}(\bm{w}) ) } = -c_i(\bm{w}) \bm{f'_i}(w_i) }$.
Now we consider all robots. Define $\mathfrak{F} \defeq \diag{\bm{f'_1}, \dots, \bm{f'_N}} \in \mbr[nN \times N]$, $K \defeq \diag{K_1, \dots, K_N} \in \mbr[nN \times nN]$ and $\bm{\Phi} \defeq \transpose{(\transpose{\bm{\Phi_1}},  \cdots, \transpose{\bm{\Phi_N}})}$. Then %
we have
\begin{align} \label{eq_phidot_multi}
	\dot{\bm{\Phi}} = - K \bm{\Phi} - \mathfrak{F} \transpose{\mathfrak{F}} K \bm{\Phi} - k_c \mathfrak{F} \bm{c(\bm{w})}.
\end{align}
One can also calculate that
$ \dot{w}_i = \matr{0 & \cdots & 0 & 1} (\bm{\vfpf_i}(\bm{\xi_i}) + k_c \bm{\vfco_i}(\bm{\xi_i})) = (-1)^n + \transpose{\bm{f'_i}(w_i)} K_i \bm{\Phi_i}(\bm{\xi_i}) + k_c c_i(\bm{w})$.
Therefore, we have that
\begin{align} \label{eq_what_dot}
	\dot{\bm{\tilde{w}}} = (-1)^n \bm{1} + \transpose{\mathfrak{F}} K \bm{\Phi} - k_c L \bm{\tilde{w}},
\end{align}
where $\bm{1} \in \mbr[N]$ is a vector consisting of all ones. 
The Laplacian matrix can be re-written as $L = D \transpose{D}$, where $D \in \mbr[N \times |\set{E}|]$ is the incidence matrix. We define the \emph{composite error vector} $\bm{e} \defeq \transpose{( \transpose{\bm{\Phi}}, \transpose{(\transpose{D} \bm{\tilde{w}})} )} \in \mbr[nN+|\set{E}|]$ and the composite gain matrix to be $\mathfrak{K} \defeq \diag{K, k_c I_{|\set{E}|}} \in \mbr[(nN+|\set{E}|) \times (nN+|\set{E}|)]$, where $I_m$ is the $|\set{E}|$-by-$|\set{E}|$ identity matrix. Therefore, from \eqref{eq_coordination_stack}, \eqref{eq_phidot_multi} and \eqref{eq_what_dot}, and noting that $D^\top \bm{1} = \bm{0}$, we have the following composite error dynamics:
\begin{align} \label{eq_error_dyn1_multi}
\scalemath{0.91}{
	\dot{\bm{e}} = \matr{\dot{\bm{\Phi}} \\ \transpose{D} \dot{\bm{\tilde{w}}} } = \matr{ - K \bm{\Phi} - \mathfrak{F} \transpose{\mathfrak{F}} K \bm{\Phi} + k_c \mathfrak{F} L \bm{\tilde{w}} \\ 
		\transpose{D} \transpose{\mathfrak{F}} K\bm{\Phi} - k_c \transpose{D} L \bm{\tilde{w}}}.
}
\end{align}
\begin{theorem}
	The coordinating vector fields $\bm{\vfcb_i}$ for $i\in \mathbb{Z}_{1}^{N}$ designed by combining \eqref{eq_vfpf}, \eqref{eq_vfco}, \eqref{eq_vf_combined} and \eqref{eq_coordination} solve Problem \ref{problem1} globally in the sense that the initial states $\bm{\xi_0} \in \mbr[(n+1) \times N]$ can be chosen arbitrarily. %
\end{theorem}
\begin{proof}
	The proof is in the extended version \cite{yao2021multiext}.
\end{proof}

\begin{remark} \label{remark_collision}
	Our work's focus is the distributed control algorithm based on guiding vector fields, so collision avoidance is not elaborated here. However, our proposed approach can incorporate existing collision avoidance algorithms \cite{yao2019integrated,wang2017safety}. For example, we can modify the \emph{nominal} guiding vector field \eqref{eq_vf_combined} in a minimally invasive way using \emph{safety barrier certificates} \cite{wang2017safe,wang2017safety}. Specifically, we add an extra term $\bm{u_i^{\mathrm{col}}} \in \mbr[n+1]$ to the vector field in \eqref{eq_vf_combined}, and this term is calculated by a quadratic program $\min \norm{\bm{u_i^{\mathrm{col}}}(t)}^2$ subject to the constraints $\dot{B}_{ij}(\bm{\xi_i}, \bm{\xi_j}) \le 1 / B_{ij}(\bm{\xi_i}, \bm{\xi_j})$ for $j \ne i$ and $i,j\in \mathbb{Z}_{1}^{N}$, where $B_{ij}(\bm{\xi_i}, \bm{\xi_j})$ is a control barrier function \cite{ames2019control} (e.g., $B_{ij}(\bm{\xi_i}, \bm{\xi_j}) = 1/(\norm{\bm{p_i}-\bm{p_j}}^2 - R^2)$, where $\bm{p}_i$ is the physical position and $R$ is the safe distance between robots). The collision avoidance behavior is shown in the supplementary video, but the theoretical analysis is left for future work.
\end{remark}

\section{Controller for A Dubins-car-like model} \label{sec: fw}
If a robot's motion can be approximately modeled by the single-integrator model, then the coordinating vector field in \eqref{eq_vf_combined} can be used directly as the velocity input to the robot. For the unicycle model, one can use feedback linearization to transform it into the single-integrator model \cite{yun1992} to utilize the guiding vector field directly. However, we will design a controller for unicycles traveling at a constant speed (i.e., the Dubins-car model) without using the feedback linearization technique. Note that the control algorithm design idea in this section is applicable to robot models of which the motions are characterized by the robot's orientations, such as the car-like model and the underwater glider model \cite{Siciliano:2007:SHR:1209344}. These models (approximately) represent many different robotic systems in reality; thus, the design methodology is widely applicable.

Different from the unicycle model, which allows backward or stationary motion, we use the following Dubins-car-like 3D model that describes fixed-wing aircraft dynamics:
	\begin{align} \label{model}
	\dot{p}_{i1} = v \cos \theta_i, \;
	\dot{p}_{i2} = v \sin \theta_i , \;
	\dot{p}_{i3} = u^{z}_{i} ,  \;
	\dot{\theta_i} = u^{\theta}_{i},
	\end{align}
where $v$ is a constant airspeed, $\transpose{(p_{i1},p_{i2},p_{i3})} \in \mbr[3]$ is the position of the $i$-th aircraft's center of mass, $\theta_i$ is the yaw angle, and $u^{z}_{i}$ and $u^{\theta}_{i}$ are two control inputs to be designed. Since the essential role of a guiding vector field is to provide the desired yaw angle to guide the flight of a fixed-wing aircraft, the core idea behind the control algorithm design is to align the aircraft's flying direction with that given by the guiding vector field. To this end, as the guiding vector field has an additional coordinate, one needs to add an additional virtual coordinate $p_{i4}$ such that the aircraft's generalized position is $(p_{i1},p_{i2},p_{i3},p_{i4}) \in \mbr[4]$. Correspondingly, its generalized velocity is $(\dot{p}_{i1}, \dot{p}_{i2} ,\dot{p}_{i3}, \dot{p}_{i4}) = (v \cos\theta_i, v\sin\theta_i, u^{z}_{i}, \dot{p}_{i4})$, where $\dot{p}_{i4}$ is virtual and extra design freedom. To align the aircraft heading $v (\cos\theta_i, \sin\theta_i)$ with the counterpart of the coordinating vector field $\bm{\vfcb_i}$, we need to ``partially normalize'' the vector field $\bm{\vfcb_i}$ such that its first two entries form a vector of the same length as $v$; that is, 
$
	v \bm{\underline{\vfcb_i}} \defeq v \bm{\vfcb_i} / \sqrt{\multivfcb{i}{1}^2 + \multivfcb{i}{2}^2}.
$
Subsequently, we design the yaw angular control input $u^{\theta}_{i}$ such that the aircraft heading $v (\cos\theta_i, \sin\theta_i)$ aligns gradually with the vector formed by the first two entries of $v \bm{\underline{\vfcb_i}}$ (i.e., $v ({\underline{\vfcb_i}}{}_{1}, {\underline{\vfcb_i}}{}_{2})$). For the last two entries of the generalized velocity $(\dot{p}_{i1}, \dot{p}_{i2} ,\dot{p}_{i3}, \dot{p}_{i4})$, one can simply equate them with those of $v \bm{\underline{\vfcb_i}}$, respectively:
\begin{align}\label{eq_u34}
\scalemath{0.89}{ \dot{p}_{i3} = u^{z}_{i}= v \multivfcb{i}{3}/ \sqrt{\multivfcb{i}{1}^2 + \multivfcb{i}{2}^2}, \quad \dot{p}_{i4} = v \multivfcb{i}{4} / \sqrt{\multivfcb{i}{1}^2 + \multivfcb{i}{2}^2}. }
\end{align}
The control algorithm design method \cite{yao2021singularity,rezende2018robust} is extended here to handle the issue with the actuator saturation in the yaw control input $u^{\theta}_{i}$ as described below. First, we define the saturation function $\sat: \mbr[] \to \mbr[]$ by $\sat(x)=x$ for $x \in [a,b]$, $\sat(x)=a$ for $x \in (-\infty, a)$ and $\sat(x)=b$ for $x \in (b, \infty)$, where $a,b \in \mbr[], a<b$ are some constants. The saturation function $\sat$ is Lipschitz continuous. For convenience, we call the time interval when $\sat(x(t))=b$ the \emph{upper saturation period}, and the time interval when $\sat(x(t))=a$ the \emph{lower saturation period}. We use $\normv{\bm{v}}$ to denote the normalization of a vector $\bm{v}$ (i.e., $\normv{\bm{v}}=\bm{v} / \norm{\bm{v}}$). We also define $\bm{\vfcb^p_i} = \transpose{(\multihatvfcb{i}{1}, \multihatvfcb{i}{2})}$,
which is the vector formed by the first two entries of the normalized vector field $\normv{\bm{\vfcb_{i}}}$. One can easily calculate that $\normv{\bm{\vfcb_i}^p} = ({\underline{\vfcb_i}}{}_{1}, {\underline{\vfcb_i}}{}_{2})$. Therefore, $\normv{\bm{\vfcb_i}^p}$ represents the orientation given by the vector field $\bm{\vfcb_i}$.

Suppose we are given 3D physical desired paths $\phy{\set{P}}_i \subseteq \mbr[3]$ parameterized by \eqref{eq_param_func}, and the coordinating vector field $\bm{\vfcb_i}: \mbr[3+N] \to \mbr[3+1]$ in \eqref{eq_vf_combined}. We denote $\bm{h_i} = \normv{\bm{h_i}}=\transpose{(\cos \theta_i, \sin \theta_i)}$ as the orientation of the aircraft,  the (signed) angle difference directed from $\normv{\bm{\vfcb^p_i}}$ to $\normv{\bm{h_i}}$ by $\sigma_i \in (-\pi, \pi]$ and define the rotation matrix $E=\left[\begin{smallmatrix}0 & -1 \\ 1 & 0\end{smallmatrix}\right]$. Then after introducing another natural assumption, one can reach Theorem \ref{thm: guidance} showing that the aircraft's orientation $\normv{\bm{h_i}}$ will converge to that of the vector field $\normv{\bm{\vfcb_i}^p}$ asymptotically (i.e., $\sigma_i$ converges to zero).
\begin{assump}
	The desired paths can be traveled by robots with saturated angular velocity, and the eventual coordination is feasible \cite{colombo2019motion}.
\end{assump}
This assumption is reasonable. It restricts the design of the desired path and the reference state vector $\bm{\omega}^*$ (hence $\bm{\Delta}^*$) to guarantee the feasibility of the robot motions, and avoid the collision of robots in the steady-state, respectively. For example, if the parametric desired path is periodic and closed, then one can simply choose $\Delta_{ij}=T/N$, where $T$ is the period, for any two neighboring indices $i,j \in \mathbb{Z}_{1}^{N}$. 
\begin{figure}[tb]
	\centering
	\subfigure{
		\includegraphics[width=0.7\linewidth]{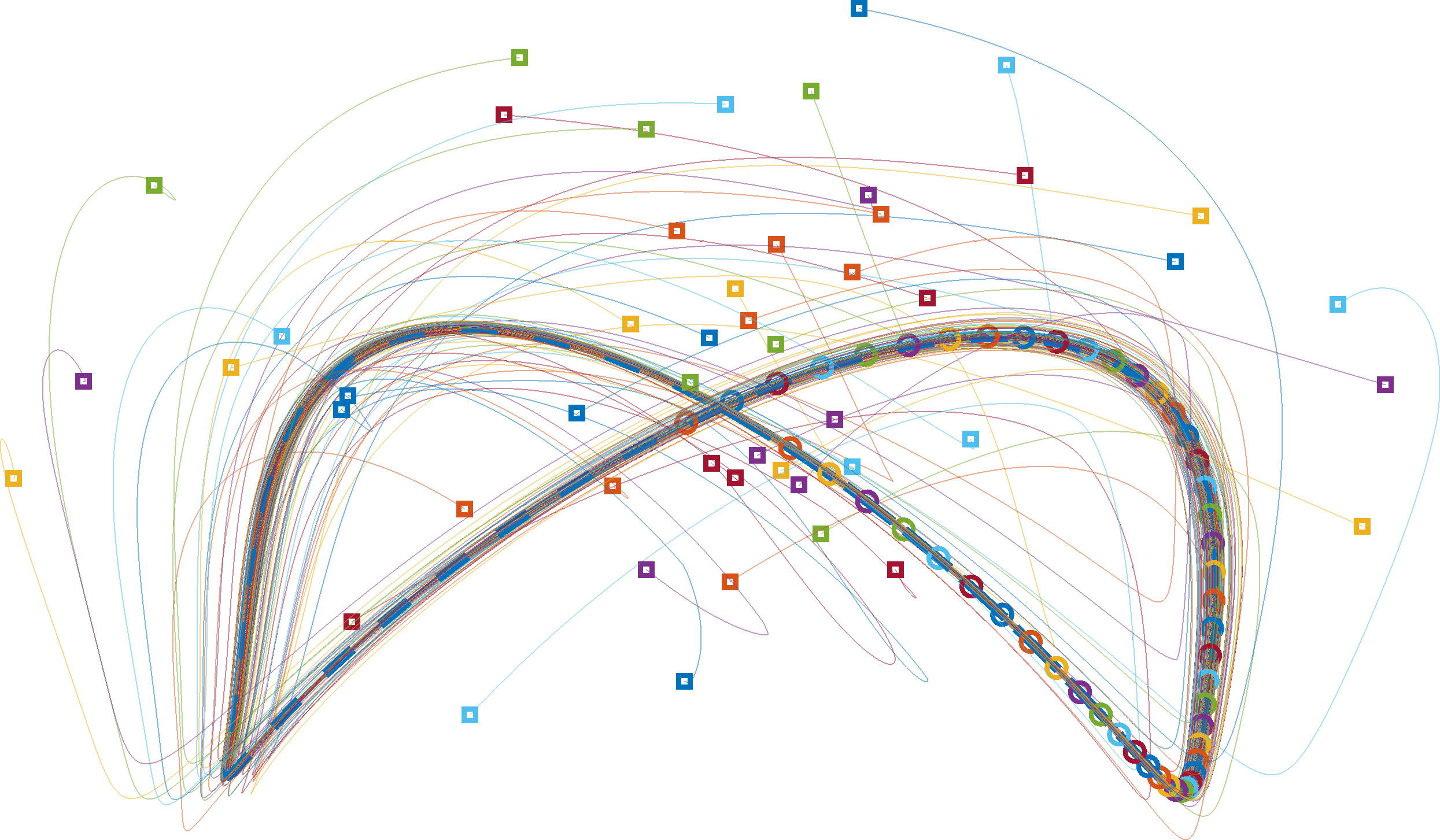}
		\label{fig:h22010200607n3n49trj}
	} \\
	\subfigure{
		\includegraphics[width=0.4\columnwidth]{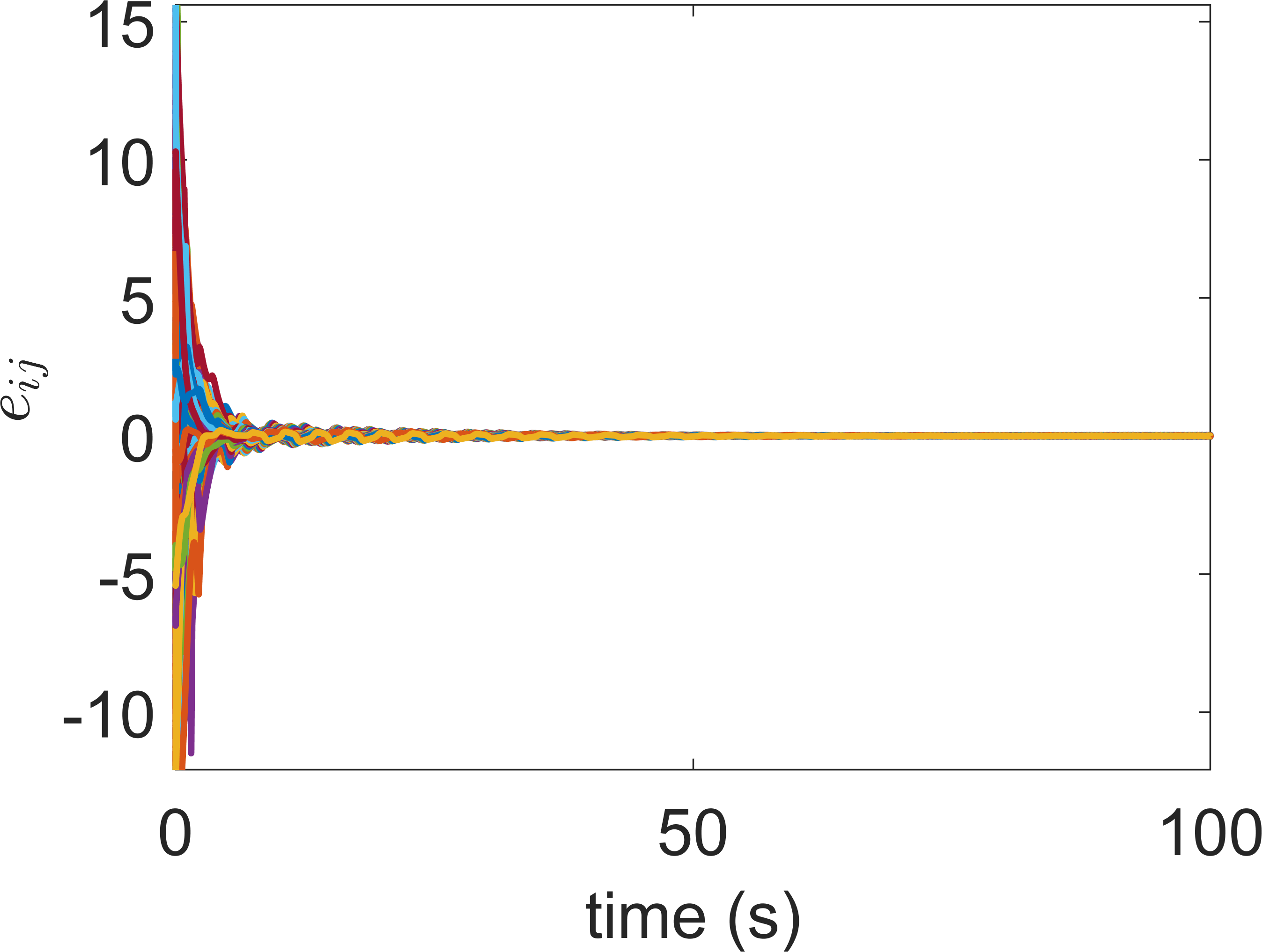}
		\label{fig:h2_201020_0955_n3N50error}
	}\hspace{-0.8em}
	\subfigure{
		\includegraphics[width=0.4\columnwidth]{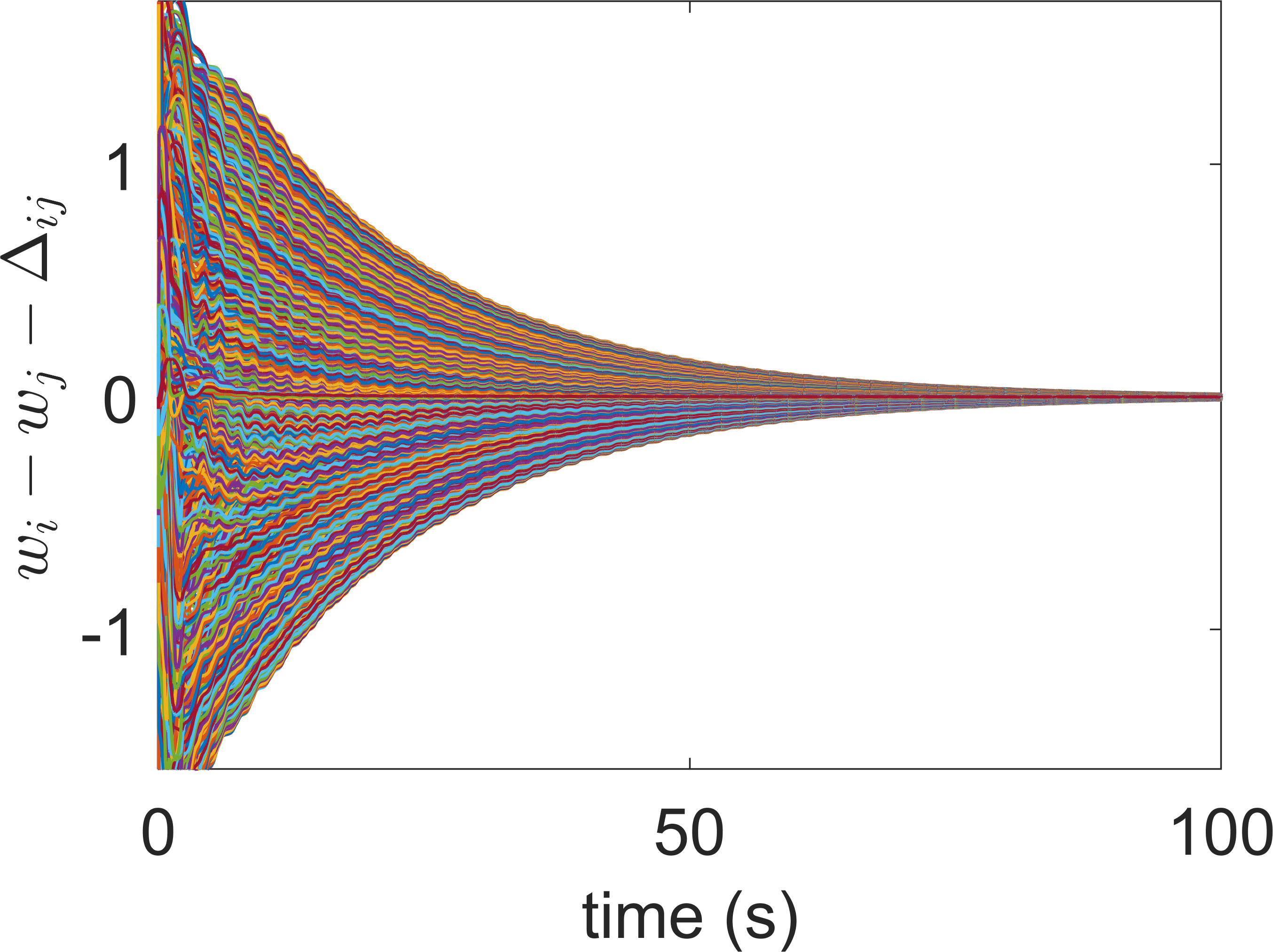}
		\label{fig:h2_201020_0955_n3N50w}
	}
	\caption{The first simulation results. The path is parameterized by $x_{i1}=15 \sin(2 w_i)$, $x_{i2}=30 \sin(w_i) \sqrt{ 0.5 (1 - 0.5 \sin^2(w_i)) }$ and $x_{i3}=5 + 5 \cos(2 w_i) - 2$ for all $i\in \mathbb{Z}_{1}^{N}$. The period of this closed curve is $T=2 \pi$ and the desired differences between two adjacent robots' virtual coordinates are $T/(2N)$. The desired relative states $\Delta_{ij}$ are constructed from the reference $w_i^*=(i-1) T/(2N)$ for $i\in \mathbb{Z}_{1}^{N}$.  The control gains are: $k_{i1}=k_{i2}=k_{i3}=1,  k_c=300$ for $i \in \mathbb{Z}_{1}^{N}$, where $k_c$ is large to accelerate the motion coordination. On top, the trajectories of robots, where squares and circles symbolize the trajectories' initial and final positions, respectively. On the bottom left, the path-following errors $e_{ij}$ for $i\in \mathbb{Z}_{1}^{N}$ and $j\in \mathbb{Z}_{1}^{n}$. On the bottom right, the coordination error $w_i-w_j-\Delta_{ij}$ for $i,j\in \mathbb{Z}_{1}^{N},i<j$. }
	\label{fig:sim1}
	\vspace{-1.6em}
\end{figure}
\begin{theorem}	\label{thm: guidance}
	Assume that the vector field satisfies $\multivfcb{i}{1}(\bm{\xi_i})^2 + \multivfcb{i}{2}(\bm{\xi_i})^2 > \gamma>0$ for $\bm{\xi_i} \in \mbr[3+1]$, $i\in \mathbb{Z}_{1}^{N}$, where $\gamma$ is a positive constant.  Let the angular velocity control input $u^{\theta}_i$  in model \eqref{model} be
	$
		\dot{\theta_i}=u^{\theta}_i = \sat( \dot{\theta}^d_i - k_\theta \transpose{\normv{\bm{h_i}}} E \normv{\bm{\vfcb^p_i}} ), \label{eq_u_theta} 
	$
	where $\dot{\theta}^d_i = -\transpose{\normv{\bm{\vfcb^p_i}}} E \bm{\dot{\vfcb}^p_i} / \norm{\bm{\vfcb^p_i}}$, $k_\theta>0$ is a constant, and $a<0$, $b>0$ are constants for the saturation function $\sat$. If the angle difference $\sigma_i$ satisfies the following conditions:
	\begin{enumerate}[leftmargin=*]
		\item The initial angle difference $\sigma_i(t=0) \ne \pi$;
		\item \label{cond2} $\sigma_i(t) \in [0, \pi)$ during the upper saturation period,  and $\sigma_i(t) \in (-\pi, 0]$ during the lower saturation period,
	\end{enumerate}
	then $\sigma_i$ will vanish asymptotically (i.e., $\sigma_i(t) \to 0$). 
\end{theorem}
\begin{proof}
	The proof is in the extended version \cite{yao2021multiext}.
\end{proof}

Although Condition \ref{cond2} in Theorem \ref{thm: guidance} might be difficult to check in practice, \emph{it conveys the core message that the saturation, albeit allowed, should not last for a long time}. However, this condition is only sufficient, while in practice, violating this condition does not immediately entail instability of the algorithm. As the aircraft is guided by the vector field, the aircraft can re-orient its heading towards the desired path even if it temporarily deviates from the desired path due to saturation or other constraints (such as path curvature), as long as the \emph{vector field orientation change rate}\footnote{As $\dot{\normv{\bm{\vfcb^p_i}}} = \dot{\theta}^d_i E \normv{\bm{\vfcb^p_i}}$, and $E \normv{\bm{\vfcb^p_i}}$ is orthogonal to $\normv{\bm{\vfcb^p_i}}$, the term $\dot{\theta}^d_i$ in \eqref{eq_u_theta} encodes how fast the vector field changes its orientation along the trajectory of the aircraft (i.e., $\dot{\theta}^d_i$ is the \emph{vector field orientation change rate}).} $\dot{\theta}^d_i$ does not saturate the control input persistently. The subsequent fixed-wing aircraft experiment verifies the effectiveness of the control law.
Nevertheless, we can remove Condition \ref{cond2} in Theorem \ref{thm: guidance} by imposing an upper bound on the magnitude of $\dot{\theta}^d_i$. Due to the page limit, the detail is in the extended version \cite{yao2021multiext}. To reduce $|\dot{\theta}^d_i|$ and avoid possible saturation, one can scale down the path parameter in the parametric functions in \eqref{eq_param_func} (e.g. by changing $\multif{i}{j}(w_i)$ to $\multif{i}{j}(\beta w_i)$, where $0<\beta<1$), or choose another desired path with a smaller curvature. Another solution is to impose an additional constraint $|u^\theta_i| \le \min\{|a|, |b|\}$ on the quadratic program mentioned in Remark \ref{remark_collision}.
\begin{remark}
	Theorem \ref{thm: guidance} guarantees that each robot's moving direction will align with the vector field eventually. Therefore, it can be shown that the convergence results of the robots' trajectories are the same as the integral curves of the guiding vector field in \eqref{eq_ode}. The rigorous theoretical analysis relies on the input-to-state stability (ISS) property \cite[Chapter 4.9]{khalil2002nonlinear}, and is left for future work.
\end{remark}

\section{Simulations and Experiments}
\subsection{Simulations}
\begin{figure}[tb]
	\centering
	\subfigure{
		\includegraphics[width=0.3\columnwidth]{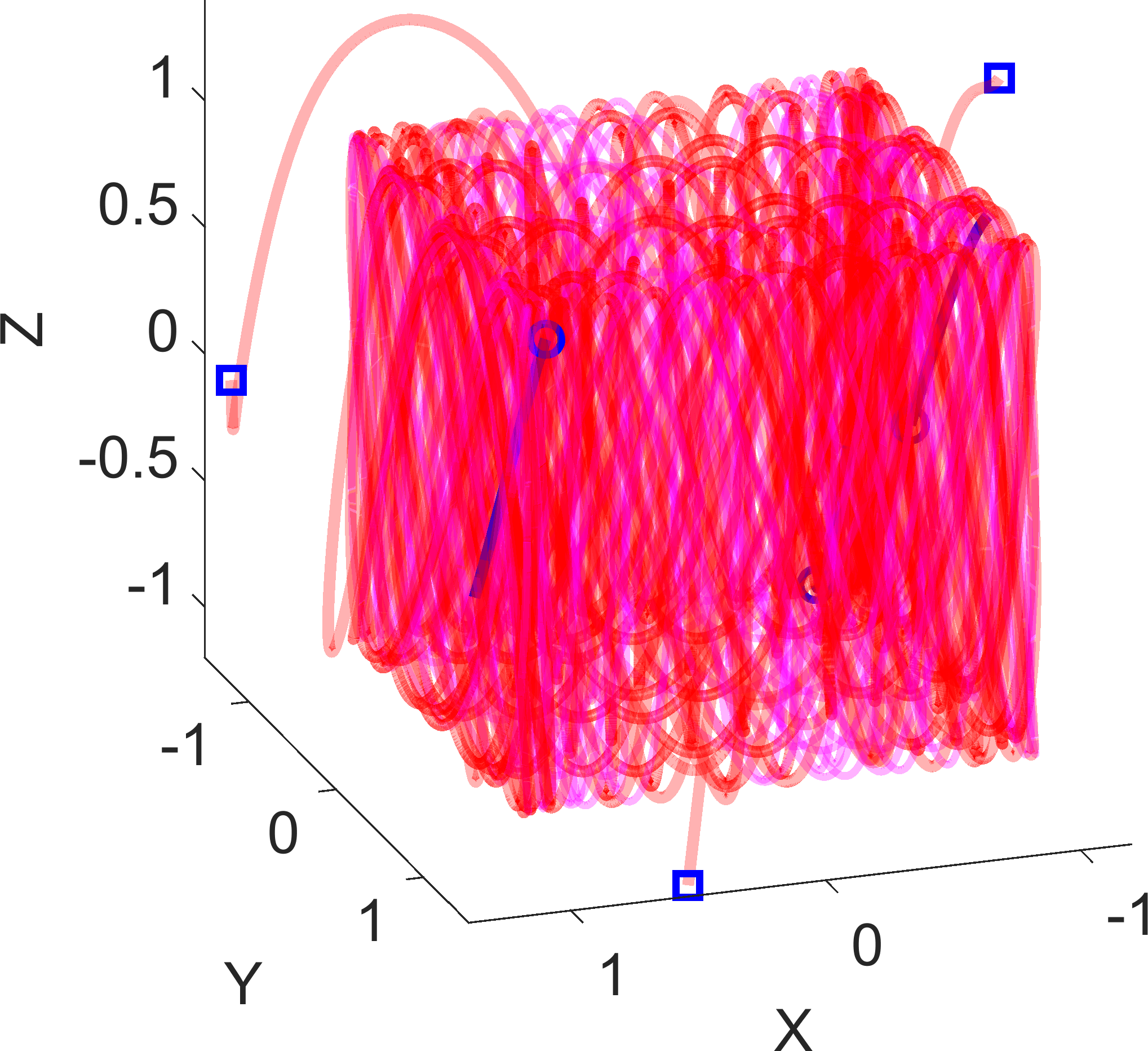}
		\label{fig:h3_201020_1550_n3N3_trj}
	}  \hspace{-1.2em}
	\subfigure{
		\includegraphics[width=0.33\columnwidth]{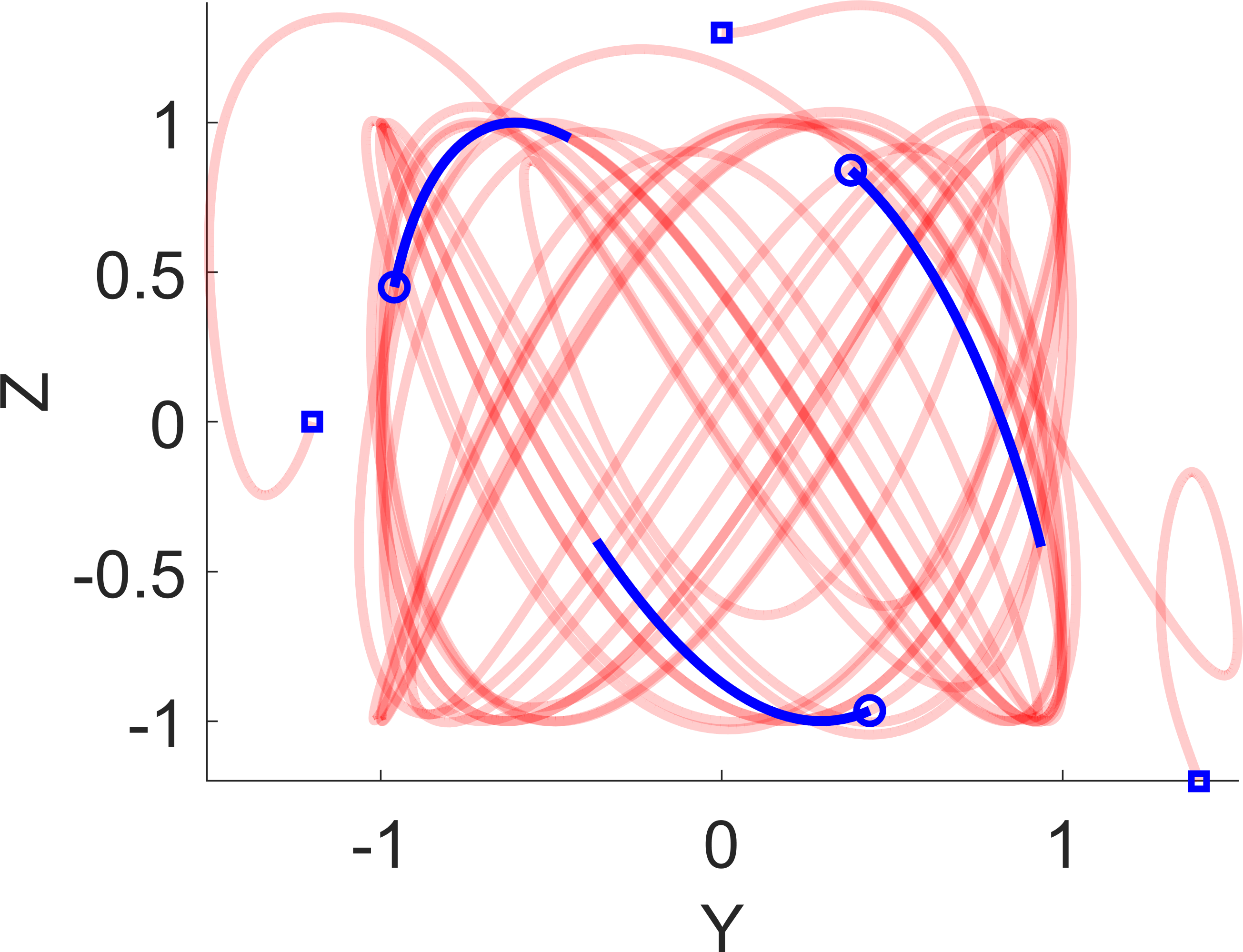}
		\label{fig:h3_201020_1550_n3N3_yz9-4}
	} \hspace{-1.2em}
	\subfigure{
		\includegraphics[width=0.33\columnwidth]{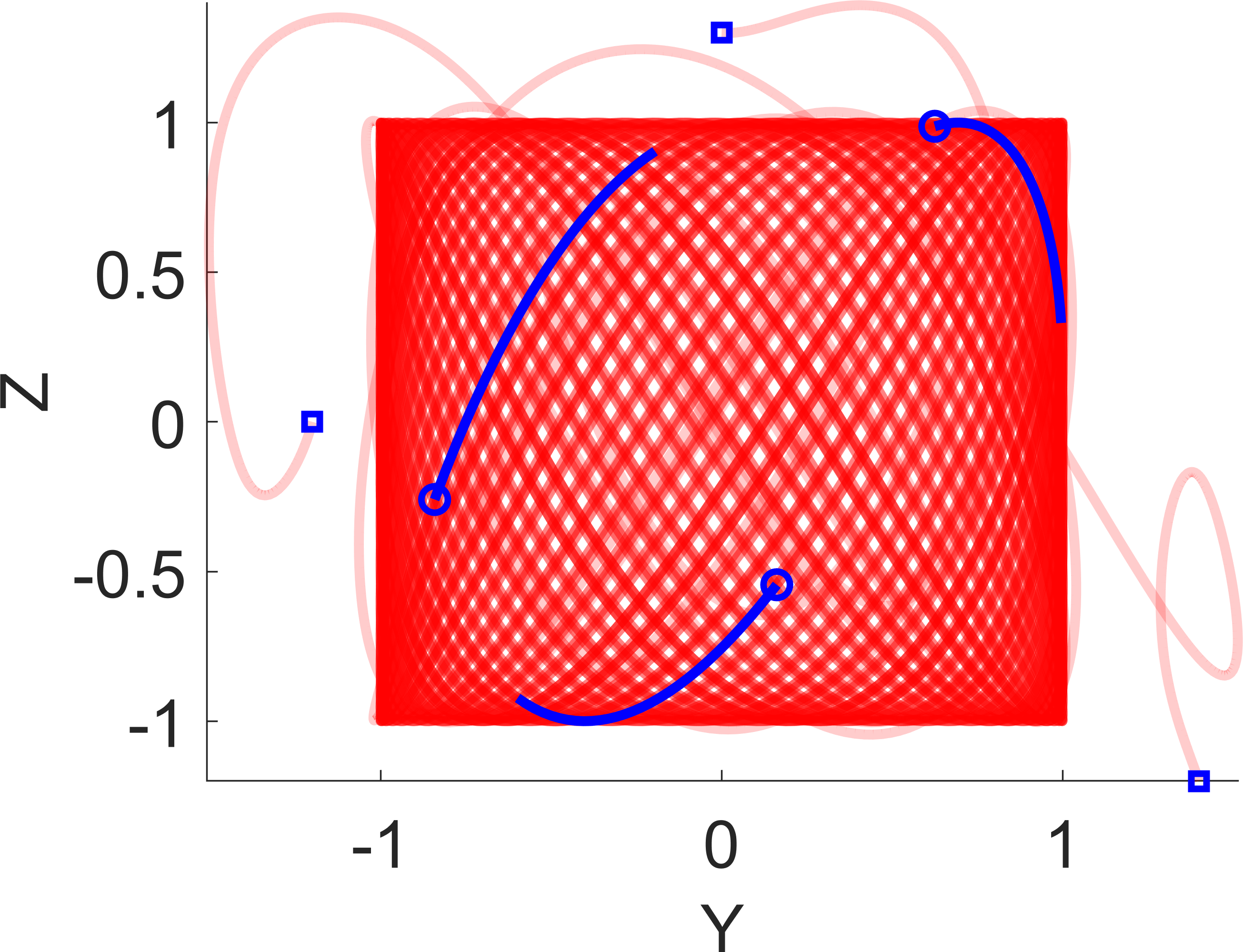}
		\label{fig:h3_201020_1550_n3N3_yz76-8}
	}
	\caption{The second simulation results. We choose a 3D Lissajous curve with \emph{irrational} coefficients: $x_{i1}=\cos(n_x w_i) + m_x$, $x_{i2}=\cos(n_y w_i) + m_y$ and $x_{i3}=\cos(n_z w_i) + m_z$ for  $i\in \mathbb{Z}_{1}^{N}$, where $n_x=\sqrt{2}, n_y=4.1, n_z=7.1, m_x=0.1, m_y=0.7, m_z=0$. This is an open curve bounded in a cube (as $n_x$ is irrational). Therefore, it is ideal for a volume coverage task.  The control gains for the coordinating vector field are: $k_{i1}=k_{i2}=k_{i3}= k_c=1$ for $i \in \mathbb{Z}_{1}^{N}$. Squares and circles symbolize trajectories' initial and final positions, and the solid blue lines are the trajectories during the last $30$ time steps. The first sub-figure shows the trajectories of three robots (red curves). The magenta curve represents part of the Lissajous curve. The last two sub-figures correspond to the Y-Z side views of the trajectories for $9.4$ and $76.8$ seconds, respectively. }
	\label{fig:sim2_1}
	\vspace{-0.6em}
\end{figure}

\begin{figure}[tb]
	\centering
	\includegraphics[width=0.5\linewidth]{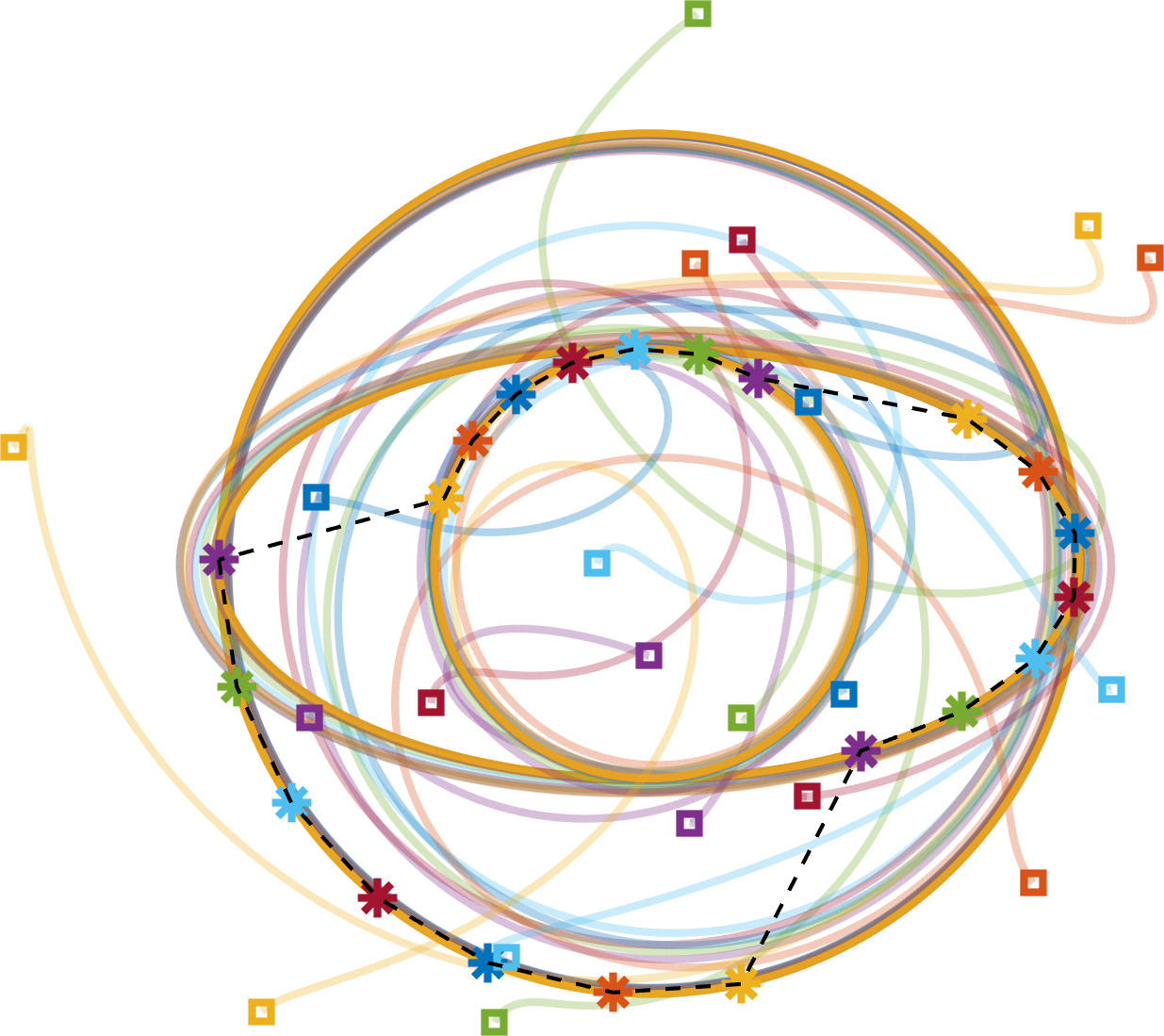}
	\caption{The third simulation results. We let $N=21$ robots follow three different paths, where the first seven robots follow a large circle of radius $a=10$, the last seven robots follow a small circle of radius $b=5$, and the remaining seven robots follow an ellipse with the semimajor axis $a$ and the semiminor axis $b$. The robots are distributed evenly in the path through the parameter $w_i$; i.e., we construct the desired relative states $\Delta_{ij}$ from the references $w_i^*(t)=(i-1) 2\pi/N$, for $i\in \mathbb{Z}_{1}^{N}$. The control gains are: $k_{i1}=k_{i2}=1, k_c=100$ for all $i\in \mathbb{Z}_{1}^{N}$, where the large value is chosen for $k_c$ to accelerate the motion coordination. The squares and $*$ symbolize the trajectories' initial and final positions, respectively. The dashed black line shows the communication links between robots and the resulting formation shape. During the steady-state, these robots do not collide into one another, since the distributed coordination guarantees that the adjacent neighbors satisfy $|w_i - w_j|=2\pi/N$, where $i$ and $j$ are any neighboring indices. }
	\label{fig:sim3}
	\vspace{-1.5em}
\end{figure}

We use \emph{cycle graphs} as the communication topology, and thus each robot is only allowed to communicate with its two adjacent neighbors. This simple communication graph results in very low communication and computation load. %
In the first simulation, we let $N=50$ robots follow a 3D \emph{self-intersecting} bent ``$\infty$''-shaped curve. As shown in Fig. \ref{fig:sim1}, all robots follow the ``$\infty$''-shaped path successfully and keep desired positions (in terms of $w_i$) between each other. 
In the second simulation, we employ $N=3$ robots to show that our algorithm is applicable for complicated and \emph{open} curves (i.e., aperiodic curve), and demonstrate its potential application to volume coverage in 3D (see Fig. \ref{fig:sim2_1}). 
In the third simulation, we show how different robots can follow different desired paths while still coordinating their motions to form some (time-varying) formation shapes (see Fig. \ref{fig:sim3}). 

\subsection{Flights with fixed-wing aircraft}
\begin{figure}[tb]
	\centering
	\includegraphics[width=1\columnwidth]{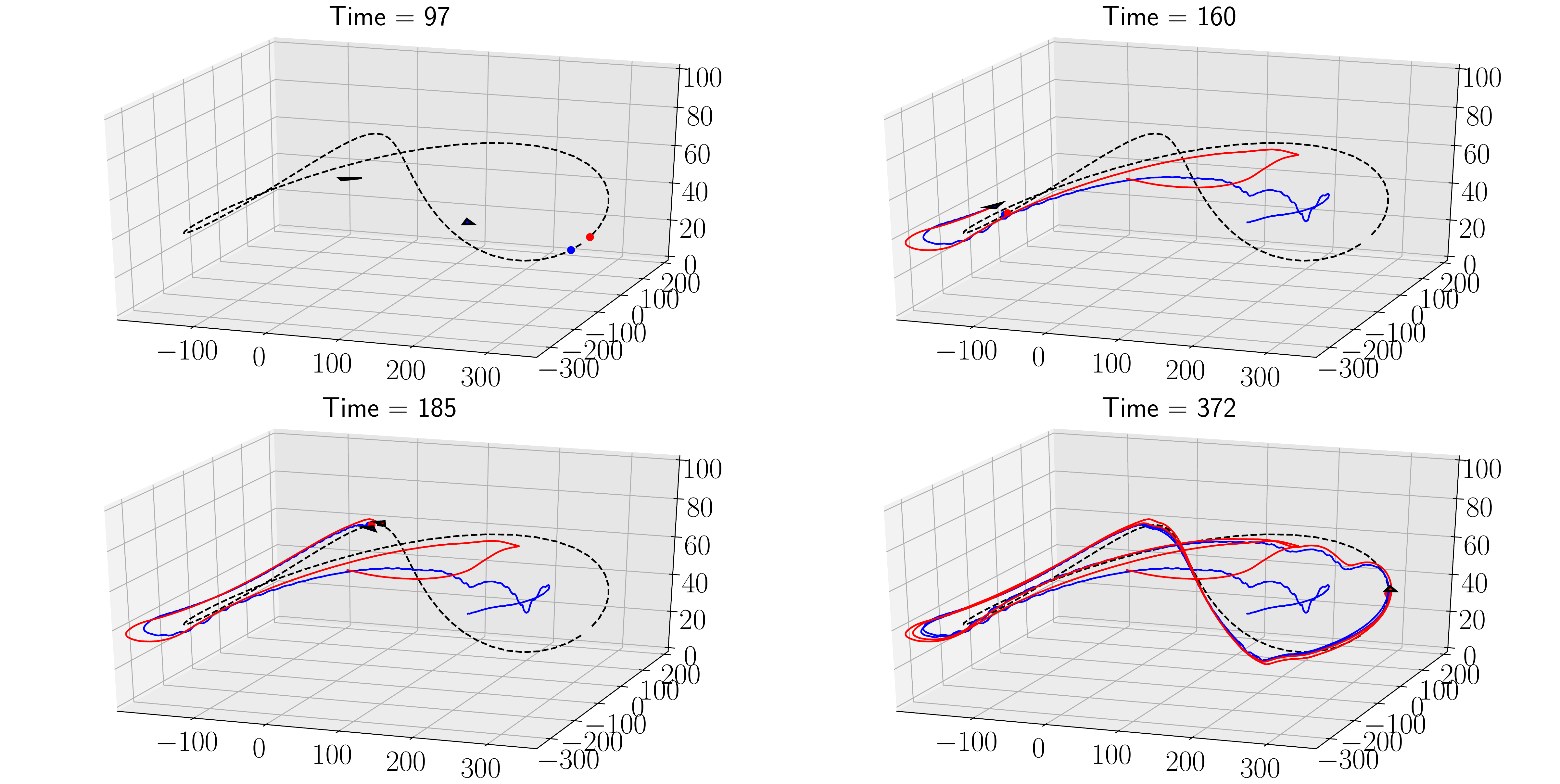}
	\caption{Two aircraft (blue and red trajectories) flying together tracking a 3D bent ``$\infty$''-shaped path. Although we show that the positions are overlapped, the same position corresponds to two different ones in reality since their GPS receptors are biased with respect to each other by around 1 meter in the XY plane.}
	\label{fig: lissaplane}
	\vspace{-0.5em}
\end{figure}

\begin{figure}[tb]
	\centering
	\includegraphics[width=1\columnwidth]{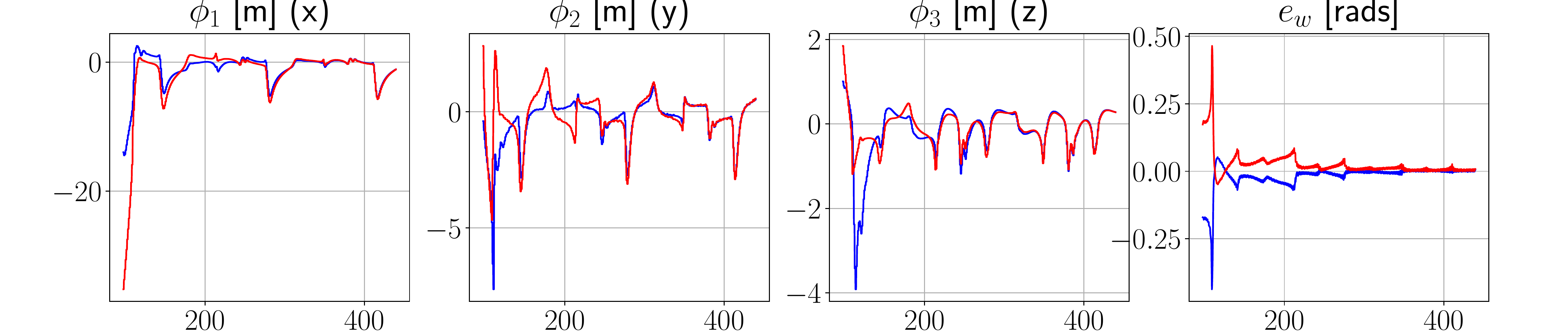}
	\caption{Distance errors (in XYZ) of the two aircraft (blue and red) to the desired 3D path, and errors with respect to the desired $\Delta w = 0$. The horizontal axes denote time in seconds.}
	\label{fig: planesphi}
	\vspace{-1em}
\end{figure}

In this experiment, two autonomous fixed-wing aircraft (i.e., Autonomous Opterra 1.2m) similar to \cite{de2017circular,kapitanyuk2017guiding2} are employed to validate Theorem \ref{thm: guidance}. The aircraft are equipped with the open-source software/hardware project \emph{Paparazzi}\cite{gati2013open}. The codes related to the proposed algorithm are in \cite{PaparaziCode20}. We choose the following 3D Lissajous curve $f_1(w) = 225 \cos(w), f_2(w) = 225\cos(2 w + \pi/2), f_3(w)= -20\cos(2 w)$, which is a bent ``$\infty$''-shaped  path. The mission requires both aircraft to have $\Delta_{12} = 0$; i.e., to fly in a tight formation. The collision is avoided by biasing the GPS measurement of one aircraft by a constant distance of $1m$ in the horizontal plane; i.e., when the aircraft achieve $\Delta_{12} = 0$, they are displaced physically. We choose $k_1 = k_2 = 0.002, k_3 = 0.0025$, $k_c=0.01, k_\theta = 1$, and the communication frequency is $10$ Hz. In the experiment, the weather forecast reported 14 degrees Celsius and a South wind of 10 km/h. In Figure \ref{fig: lissaplane}, the telemetry shows that both airplanes converge to fly together while following the path. The experiment shows that once an aircraft goes faster than its partner, the algorithm guides the airplane to deviate from the curve to travel more distances to ``wait'' for its partner. Nevertheless, these deviations from the desired path are within the order of one or two meters (see Figure \ref{fig: planesphi}).

Note that the employed aircraft do \emph{not} control their ground speeds. In fact, they have a reference signal in their throttle to keep a safe airspeed, and the aircraft increase/decrease such a reference to ascend/descend while following the 3D path. %
A traditional trajectory tracking algorithm would \emph{force} the aircraft to track an \emph{open-loop} $f_1(t), f_2(t)$ and $f_3(t)$; i.e., it requires controlling the airspeed/ground speed of the aircraft. Such a requirement is demanding (or unsatisfiable) if the aircraft are not equipped with the required sensors or actuators (e.g., spoilers/flaps while descending), and the wind always affects the speed of the airplane. By contrast, our guidance algorithm is free from such a requirement since the parameter $w$ is in a \emph{closed-loop} with the aircraft state, and thereby adapting automatically to the position and velocity of the aircraft.

\section{Conclusions and future work}
We design a coordinating vector field for multiple robots to follow (possibly different) desired paths while coordinating their motions distributedly, and a control algorithm is designed for the Dubins-car-like model. In particular, we use the path parameter $w_i$ as an additional virtual coordinate for the guiding vector field, which is further utilized as local information communicated among neighboring robots to achieve motion coordination. The vector field is rigorously proved to enable robots to follow their respective desired paths while coordinating their motions. Our algorithm is a significant extension of the circumnavigation algorithms and circular formation control algorithms since it enables multiple robots to cooperatively move along arbitrary paths and  form different (possibly varying) desired shapes. Moreover, our algorithm requires a very low amount of local communication. Simulations and experiments support the theoretical results.
Future work includes detailed analysis of the collision avoidance using control barrier functions, and performance evaluation in the presence of localization errors. We will compare our algorithm with the approaches mentioned in the Introduction in our future work, and will also consider the problem on smooth manifolds.
	

\end{document}